%% file: main.tex
\documentclass[10pt]{article} 
\usepackage[preprint]{tmlr}

\input{math_commands.tex}

\usepackage{hyperref}
\usepackage{url}
\usepackage{graphicx}
\usepackage{subcaption} 
\usepackage{booktabs}
\usepackage{makecell}
\usepackage{multirow}
\usepackage{multicol}

\title{\textbf{MONKEY}: \textbf{M}asking \textbf{ON KEY}-Value Activation Adapter for Personalization }

\author{\name James Baker \email jbaker15@umbc.edu \\
      \addr University of Maryland, Baltimore County
      }

\def\month{MM}  
\def\year{YYYY} 
\def\openreview{\url{https://openreview.net/forum?id=XXXX}} 

\begin{document}

\maketitle

\begin{abstract}
Personalizing diffusion models allows users to generate new images that incorporate a given subject, allowing more control than a text prompt. These models often suffer somewhat when they end up just recreating the subject image and ignoring the text prompt. We observe that one popular method for personalization, IP-Adapter, automatically generates masks that segment the subject from the background during inference. We propose to use this automatically generated mask on a second pass to mask the image tokens, thus restricting them to the subject, not the background, allowing the text prompt to attend to the rest of the image. For text prompts describing locations and places, this produces images that accurately depict the subject while definitively matching the prompt. We compare our method to a few other test time personalization methods, and find our method displays high prompt and source image alignment. We also perform a user study to validate whether end users would appreciate our method. Code available at \url{https://github.com/jamesBaker361/monkey}
\end{abstract}
\section{Introduction}

Diffusion models  have become a dominant paradigm for image generation \citep{ho2020denoisingdiffusionprobabilisticmodels,rombach2022highresolutionimagesynthesislatent}, competitive with autoregressive models \citep{parmar2018imagetransformer,xiong2025autoregressivemodelsvisionsurvey}, and largely superseding GANs \citep{goodfellow2014generativeadversarialnetworks}. However, despite their expressive power, end users often desire more control over who or what appears in the generated image. Personalization methods address this need by allowing a model to incorporate a specific person or object into new generations while retaining the flexibility of text-based conditioning and adhering to the text prompt.

Existing personalization methods fall into the categories of test-time fine tuning, such as Dreambooth \citep{ruiz2023dreamboothfinetuningtexttoimage} and Textual Inversion \citep{gal2022imageworthwordpersonalizing}, or adapter-based methods like IP-Adapter \citep{ye2023ipadaptertextcompatibleimage} and InstantID \citep{wang2024instantidzeroshotidentitypreservinggeneration}. Both categories often struggle to balance fidelity to the subject with faithfulness to the text prompt. In particular, when strong visual features of the subject dominate, the model tends to reproduce the original subject image rather than composing it naturally into new scenes or contexts. This results in limited generalization and weaker prompt alignment.

We observe that one popular personalization method—IP-Adapter—implicitly performs segmentation. Specifically, the adapter produces attention maps in the intermediate UNet transformer blocks, which attend to semantic content features, that effectively separate the subject from the background. This implies that UNets combined with a pretrained adapter already contain sufficient information to localize the subject without explicit supervision. Given this, we propose  obtaining the implicit subject mask from IP-Adapter, which we apply in a second pass to mask the image tokens so that they only attend to the subject region. This ensures that the text prompt can guide generation in the background, allowing higher text alignment while still preserving the subject features. Our contributions are:
\begin{itemize}
    \item We identify how different IP Adapter tokens at different layers correspond to the subject and background of generated images
    \item We use this for a two-stage inference process, where we use the first stage of inference to generate the mask, and then regenerate the image using the mask on the relevant image tokens to better align the background with the text prompt. 
    \item Our method, which we call MONKEY Adapter, requires training no new weights or additional modules, and either outperforms other adapter-based personalization methods or exists on the pareto frontier of text and subject alignment.
\end{itemize}
\section{Related Work}

\subsection{Personalization}
Text-to-image diffusion models \citet{Croitoru_2023,chen2025comprehensive} have been incredibly impressive in their ability to generate diverse, realistic images. While generally prompted on just text, a diffusion model can also be prompted on a particular instance of a category. For example, a user may want to generate pictures of a specific cat or person, instead of a generic cat or person. This is the task of Personalization. A user should be able to supply a few, or even a single image of a concept (such as a face, object or person) and be able to reliably generate new images of that concept that match input prompts. We follow the dichotomy of \citep{zhang2024survey} of dividing personalization methods into test-time fine tuning or pretrained adaptation. The former group requires the diffusion model to be retrained on each new concept. Some examples are Dreambooth \citep{ruiz2023dreamboothfinetuningtexttoimage}, Textual Inversion \citep{gal2022imageworthwordpersonalizing} and Imagic \citep{kawar2023imagictextbasedrealimage}. Pretrained adaptation does not require any new training. Concept images can be supplied to the diffusion model and personalized images immediately created at inference. Some examples include ID-Aligner \citep{chen2024idalignerenhancingidentitypreservingtexttoimage}, Face2Diffusion \citep{shiohara2024face2diffusionfasteditableface}, and most relevant to this work, IP-Adapter \cite{ye2023ipadaptertextcompatibleimage}. IP-Adapter embeds the concept image using CLIP \citep{radford2021learningtransferablevisualmodels} and learns an embedding for each layer of the UNet. The layer-wise embedding attends to the output of each layer using cross-attention.
\subsection{Extracting Features from UNets}
Unets \citep{ronneberger2015unetconvolutionalnetworksbiomedical} are a common backbone model for diffusion \citep{ho2020denoisingdiffusionprobabilisticmodels}, where they are trained to predict the noise from a corrupted image. At inference time, they gradually reduce pure noise to an actual image. The later layers of models like ResNet \citep{he2015deepresiduallearningimage} or VGG \citep{simonyan2015deepconvolutionalnetworkslargescale}, can be used as semantic feature extractors, while earlier layers extract stylistic information. For UNets, the center "bottleneck" layers extract  semantic content features, while the earlier layers learn style features \citep{haas2024discoveringinterpretabledirectionssemantic,xu2024freetunersubjectstyletrainingfree,frenkel2024implicitstylecontentseparationusing,schaerf2025trainingfreestylecontenttransfer}. This property has been used for personalization, such as doing text inversion using a new token for each layer for each new concept \citep{voynov2023pextendedtextualconditioning,agarwal2023imageworthmultiplewords}. Different timesteps of the inference process are also more influential than others in determining the content features of the final images \citep{yu2023freedomtrainingfreeenergyguidedconditional,zhang2023prospectpromptspectrumattributeaware,agarwal2023imageworthmultiplewords}.

\section{Methodology}
\subsection{Attention Maps}
The original IP-Adapter \cite{ye2023ipadaptertextcompatibleimage} checkpoint, for each layer of the diffusion model, produces four tokens that are concatenated to the text tokens during the cross attention stage. During cross attention, the intermediate features \(\phi_\ell\) are passed to each layer, and multiplied by the query matrix to get \(Q\) and then the text and image token embeddings are multiplied by the key and value matrices to get \(K\) and \(V\). The final output of the attention layer is \(\textrm{softmax}(\frac{QK^T}{\sqrt{d}}) \cdot V\). The \(\textrm{softmax}( \frac{QK^T}{\sqrt{d}})\) component represents the degree of correspondence between the semantic embeddings of each token and each spatial location in the image. Given that deeper levels of UNets correspond to semantic features, we decided to plot the activations of the text and ip tokens of an intermediate layer of each part of the generated images.

\begin{figure}
    \centering
    \includegraphics[width=0.75\linewidth]{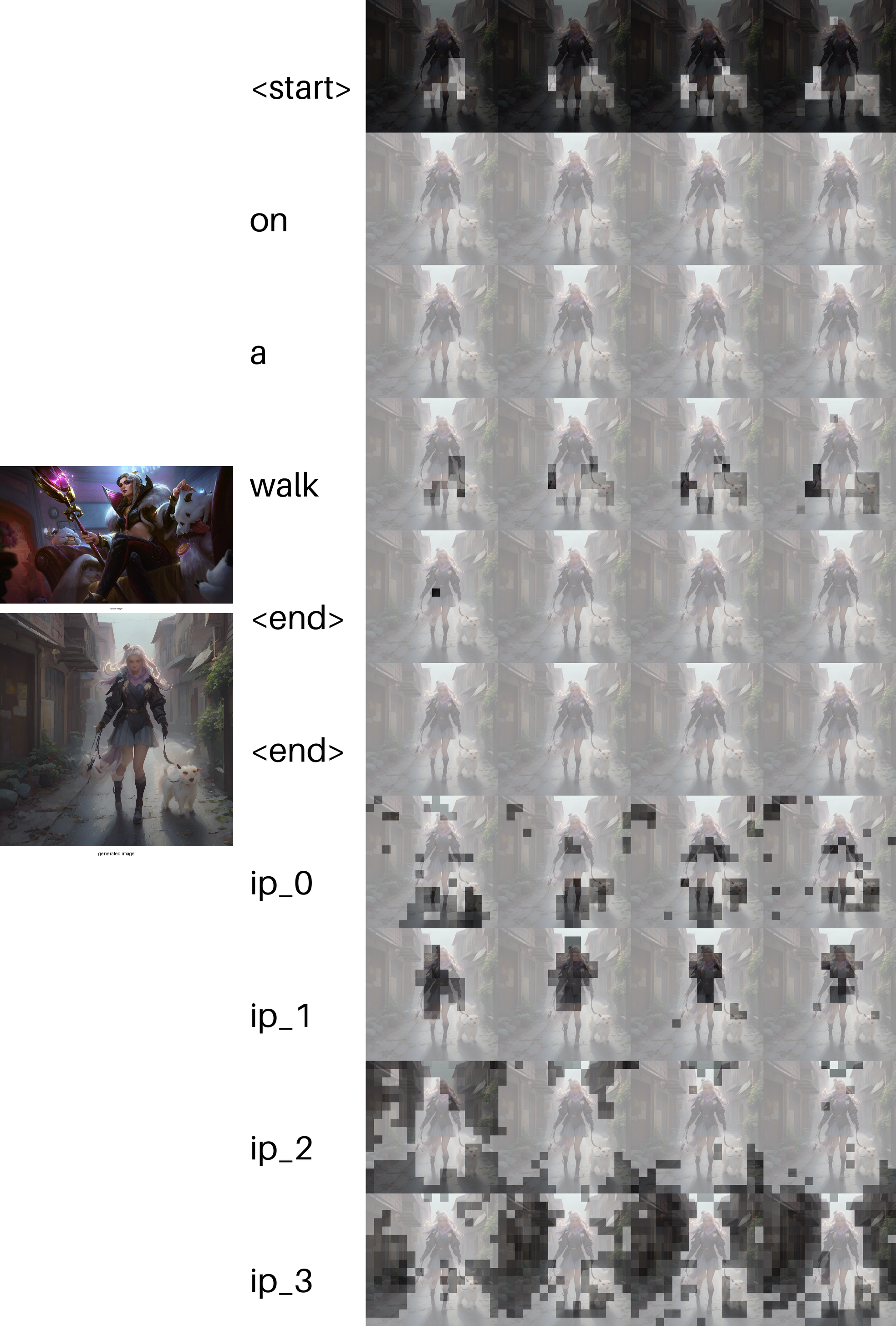}
    \caption{Token Attention Maps}
    \label{fig:maps}
\end{figure}

We use the \url{https://huggingface.co/SimianLuo/LCM_Dreamshaper_v7} checkpoint from huggingface, which is a latent consistency \citep{luo2023latentconsistencymodelssynthesizing} distillation of \url{https://huggingface.co/Lykon/dreamshaper-7}. For the IP-Adapter, we used the standard \url{https://huggingface.co/h94/IP-Adapter} checkpoint. We used the second transformer in the first up-block layer named \url{up_blocks.1.attentions.1.transformer_blocks.0.attn2}. For some visualization of the attention maps of other layers, refer to \ref{appendix:maps}

As can be seen in Figure \ref{fig:maps}, we can observe that different keys attend to different regions of the image; \(ip_1\) attends to the actual subject itself, meanwhile \(ip_2\) and \(ip_3\) attend to the background. 

\subsection{Inference}
Diffusion models themselves are deterministic. The noise they denoise to generate realistic images are randomly drawn from a gaussian normal distribution, but the same initial noise will produce the same output. Given that we can extract a mask that captures the subject, a second pass with the same initial random noise with, where we mask the ip tokens will recreate the foreground components of the original image that feature the subject, but the background will only be attended to by the text tokens. For the first image, we used 4 inference steps and averaged the masks from the second and third steps to generate the mask. Then for the final image, we used eight inference steps and applied the mask to the third to sixth steps. We call our approach the \textbf{MONKEY} Adapter, standing for \textbf{M}asking \textbf{ON KEY}-Value Activation Adapter.

\section{Experiments}
\subsection{Models}
We compared our method to a few other test-time personalization methods. Given that our method did not require training any new models or weights, we chose works that also did not train any new weights, those being FreeGraftor \citep{yao2025freegraftortrainingfreecrossimagefeature}, RectifID \citep{sun2024rectifidpersonalizingrectifiedflow}, MASA \citep{cao2023masactrltuningfreemutualselfattention} and TF-I2I \citep{hsiao2025tfti2itrainingfreetextandimagetoimagegeneration}, in addition to  the baseline ip-adapter checkpoint with the ip token scale set to 0.5 and 1.0. 
\subsection{Data}
We used two image datasets: the Dreambooth dataset \citep{ruiz2023dreamboothfinetuningtexttoimage}, consisting of objects and animals, and our own curated dataset of illustrations of characters from the card game \textit{Magic the Gathering}, which featured mostly humanoids with various fantastical features and diverse appearances. For the text prompts, we used a set of prompts relating to background and location such as \textit{on top of green grass with sunflowers around it} and \textit{on a cobblestone street}, generated by asking ChatGPT \citep{openai2023gpt4}. A complete list can be found in \ref{appendix:prompts}.
\subsection{Results}
\subsubsection{Automated Evaluation}
We used DINOv2 Identity \citep{oquab2024dinov2learningrobustvisual}, CLIP Image and CLIP Text embedding similarity. The first two measure similarity between the source and generated images, and the third measures similarity between the text prompt and generated image. Scores for each dataset are shown in Table \ref{tab:score}. We provide visual comparisons using some handpicked results in Table \ref{tab:visuals}

\begin{figure}[htbp]
  \centering
  \begin{subfigure}{0.45\textwidth}
    \centering
    \includegraphics[width=\linewidth]{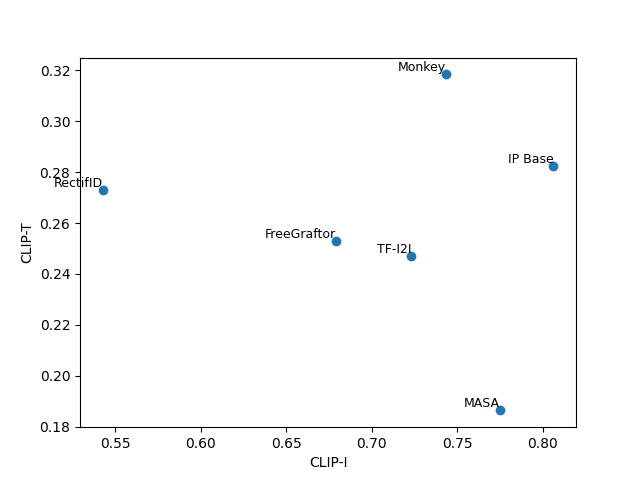}
    \caption{Dreambooth dataset}
    \label{fig:image1}
  \end{subfigure}
  \hfill
  \begin{subfigure}{0.45\textwidth}
    \centering
    \includegraphics[width=\linewidth]{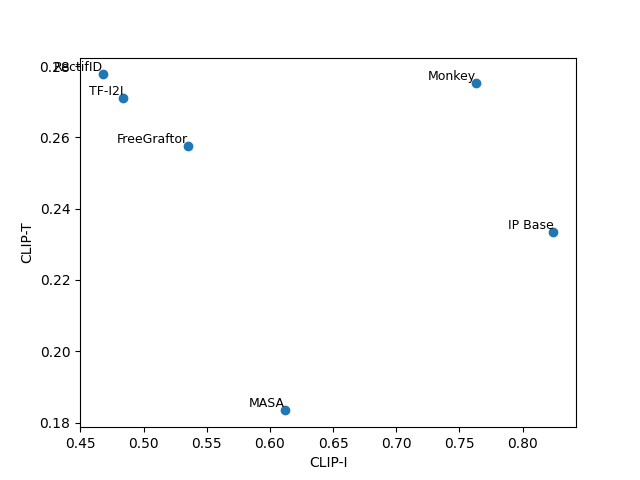}
    \caption{Magic dataset}
    \label{fig:image2}
  \end{subfigure}
  \caption{CLIP Image vs CLIP Text similarities for each method}
  \label{fig:two_images}
\end{figure}

\begin{table}[ht]
    \centering
    \begin{tabular}{|c|c|c|c|c|c|c|}
    \hline
      \multirow{2}{*}{Model}   & \multicolumn{3}{c|}{Magic } &  \multicolumn{3}{c|}{Dreambooth} \\
      
       & CLIP-T & DINO & CLIP-I & CLIP-T & DINO & CLIP-I  \\
       \hline
        TF-I2I & 0.271 & 0.091 & 0.484  & 0.247 & 0.413 & 0.723  \\ 
RectifID & \textbf{0.278} & 0.126 & 0.468  & 0.273 & 0.14 & 0.543  \\ 
MASA & 0.184 & 0.299 & 0.614  & 0.186 & 0.511 & \textit{0.775}  \\ 
FreeGraftor & 0.258 & 0.29 & 0.535  & 0.253 & 0.471 & 0.679  \\ 
IP Base & 0.233 & \textbf{0.565} & \textbf{0.824}  & \textit{0.282} & \textbf{0.621} & \textbf{0.806}  \\ 
Monkey & \textit{0.275} & \textit{0.493} & \textit{0.763}  & \textbf{0.318} & \textit{0.529} & 0.743  \\ 
\hline
    \end{tabular}
    \caption{Scores- Best results are bolded, second best are italicized}
    \label{tab:score}
\end{table}

Our method exists on the pareto frontier of other methods. For the Dreambooth dataset, it offers the highest text alignment and the third-best or second-best image alignment, using the CLIP-I and DINO metrics, respectively. For the magic dataset, it offered the second best text and image and text alignment.

\begin{table}[ht]
    \centering
    \begin{tabular}{c|c|c|c|c|c|c}
      \includegraphics[scale=0.225]{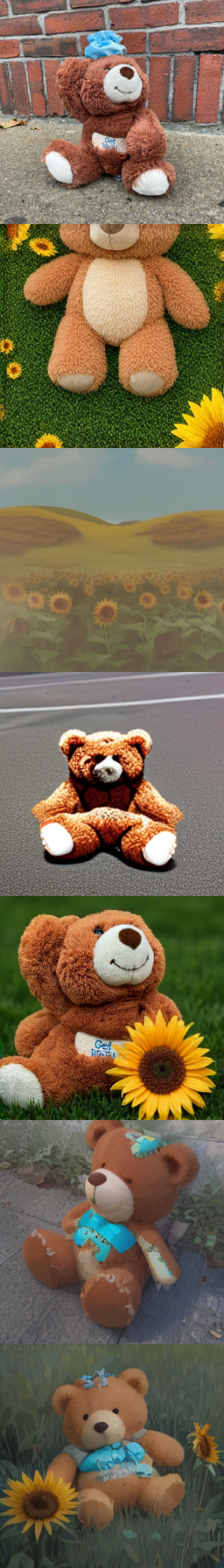}   & \includegraphics[scale=0.225]{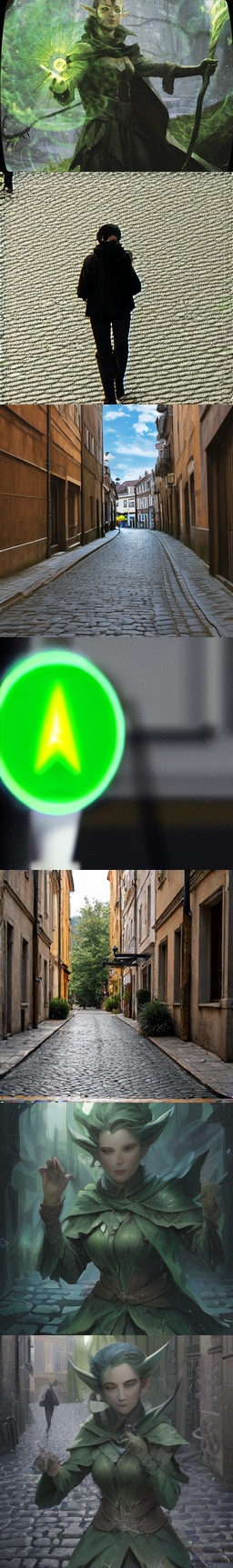} & \includegraphics[scale=0.225]{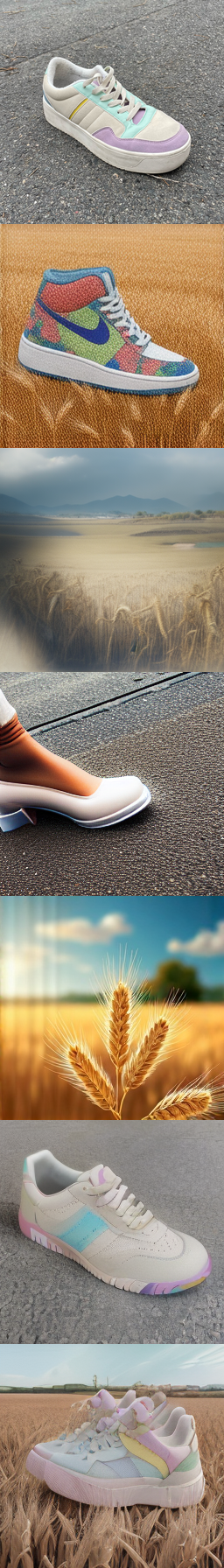} & \includegraphics[scale=0.225]{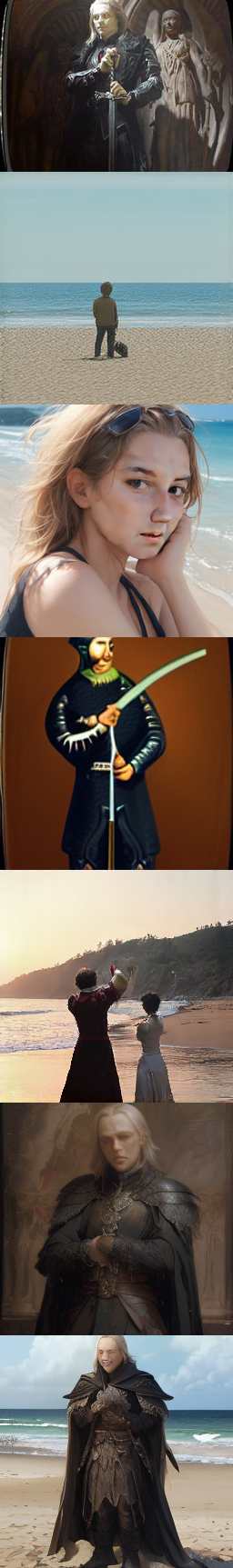} & \includegraphics[scale=0.225]{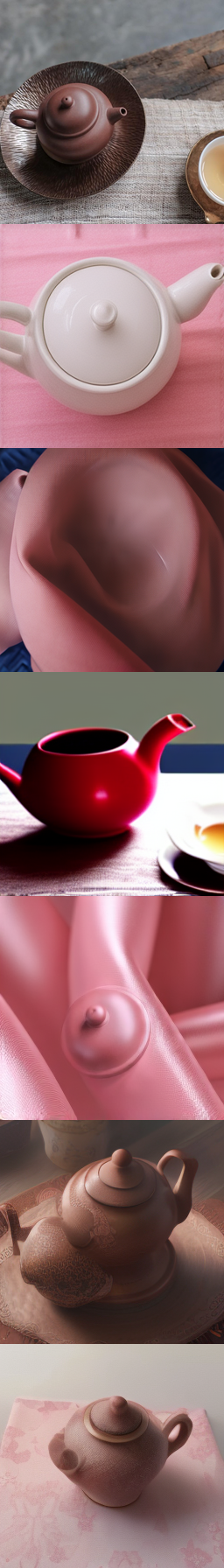} & \includegraphics[scale=0.225]{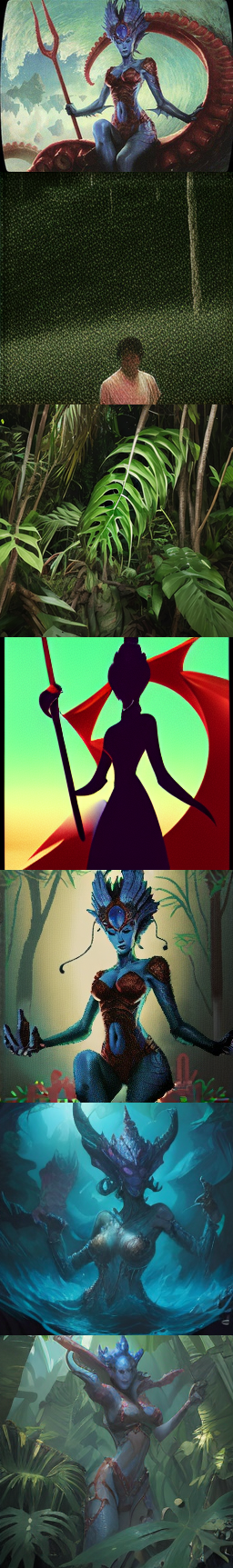} & \includegraphics[scale=0.225]{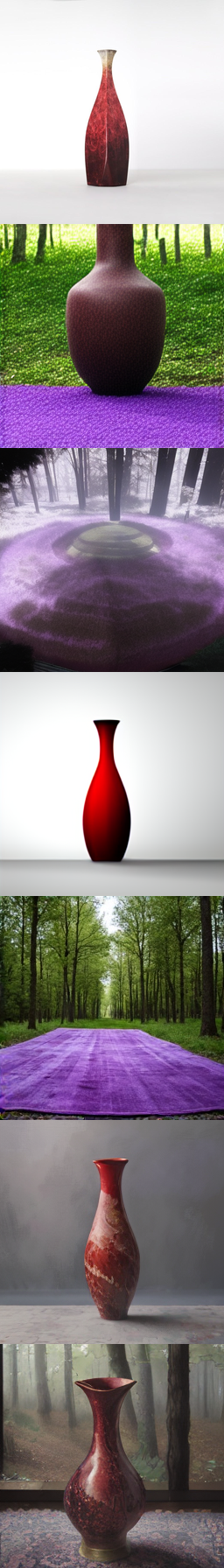} \\
         
    \end{tabular}
    \caption{Image Comparisons. Prompts used (from left to right): \textit{on top of green grass with sunflowers around it}, \textit{on a cobblestone street}, \textit{with a wheat field in the background}, \textit{on the beach}, \textit{on top of pink fabric}, \textit{in the jungle}, and \textit{on top of a purple rug in the forest}. The top image is the source image, followed by the images generated by (from second to top to bottom), \textbf{TF-I2I}, \textbf{RectifID}, \textbf{MASA}, \textbf{FreeGraftor}, \textbf{Baseline IP-Adapter}, and \textbf{MONKEY} (ours)}
    \label{tab:visuals}
\end{table}

\subsubsection{User Study}
We also performed a user study to gauge how real humans liked our methods. We used \url{prolific.com} to gather survey respondents. Each was given 40 questions, consisting of a source image and a prompt, and was asked to choose which generated image best matched the prompt and source image. We did 20 source images from each dataset. We surveyed 30 people, for a total of 1200 responses. We counted the frequency that the image produced by each image was chosen as the best. Notably, we did \textit{not} ask users to separately evaluate text and image alignment- instead we asked for both. The reason for this was that we wanted to allow users to choose the method that had the optimal balance of prompt and image alignment. Each participant was paid \(\$5.00\) for approximately ten minutes of work. Results are shown in table \ref{tab:counts}. The MONKEY method was chosen as the best a plurality of the time across both datasets. Breaking it down, we see that it was only the \textit{second} most popular choice for the dreambooth dataset. These results demonstrate that our method is well aligned with human preferences.

\begin{table}[]
    \centering
    \begin{tabular}{|c|c|c|c|c|c|c|}
    \hline
        Data & \multicolumn{6}{c|}{Method}  \\
        \hline
 & TF-I2I & RectifID & MASA & FreeGraftor & IP Adapter & MONKEY \\ 
 \hline
Magic & 15 & 2 & 35 & 86 & 118 & 344 \\ 
Dreambooth & 60 & 0 & 4 & 376 & 27 & 133 \\  
Aggregated & 75 & 2 & 39 & 462 & 145 & 477 \\ 
\hline
    \end{tabular}
    \caption{Counts}
    \label{tab:counts}
\end{table}

\newpage
\section{Conclusion and Future Work}
We presented a simple inference-time masking strategy that improves the balance between subject image alignment and text prompt alignment in adapter-based personalization of diffusion models. By leveraging the implicit subject masks already present in IP-Adapter activations, our method enhances compositional control without training any new weights or retraining pre-existing weights. Future work can extend this approach toward  multi-subject personalization and combining MONKEY Adapter with complementary personalization techniques such as ControlNet \citep{zhang2023addingconditionalcontroltexttoimage}.



\bibliography{main}
\bibliographystyle{tmlr}

\appendix
\section{Appendix}
You may include other additional sections here.
\subsection{Text Prompts}\label{appendix:prompts}
A complete list of all text prompts used:
\begin{itemize}
    \item in the jungle
\item in the snow
\item on the beach
\item on a cobblestone street
\item on top of pink fabric
\item on top of a wooden floor
\item with a city in the background
\item with a mountain in the background
\item with a blue house in the background
\item on top of a purple rug in a forest
\item with a wheat field in the background
\item with a tree and autumn leaves in the background
\item with the Eiffel Tower in the background
\item floating on top of water
\item floating in an ocean of milk
\item on top of green grass with sunflowers around it
\item on top of a mirror
\item on top of the sidewalk in a crowded street
\item on top of a dirt road
\item on top of a white rug

\end{itemize}

\subsection{Additional IP Attention Maps}\label{appendix:maps}
\newpage
\begin{figure}
    \centering
    \includegraphics[width=0.75\linewidth]{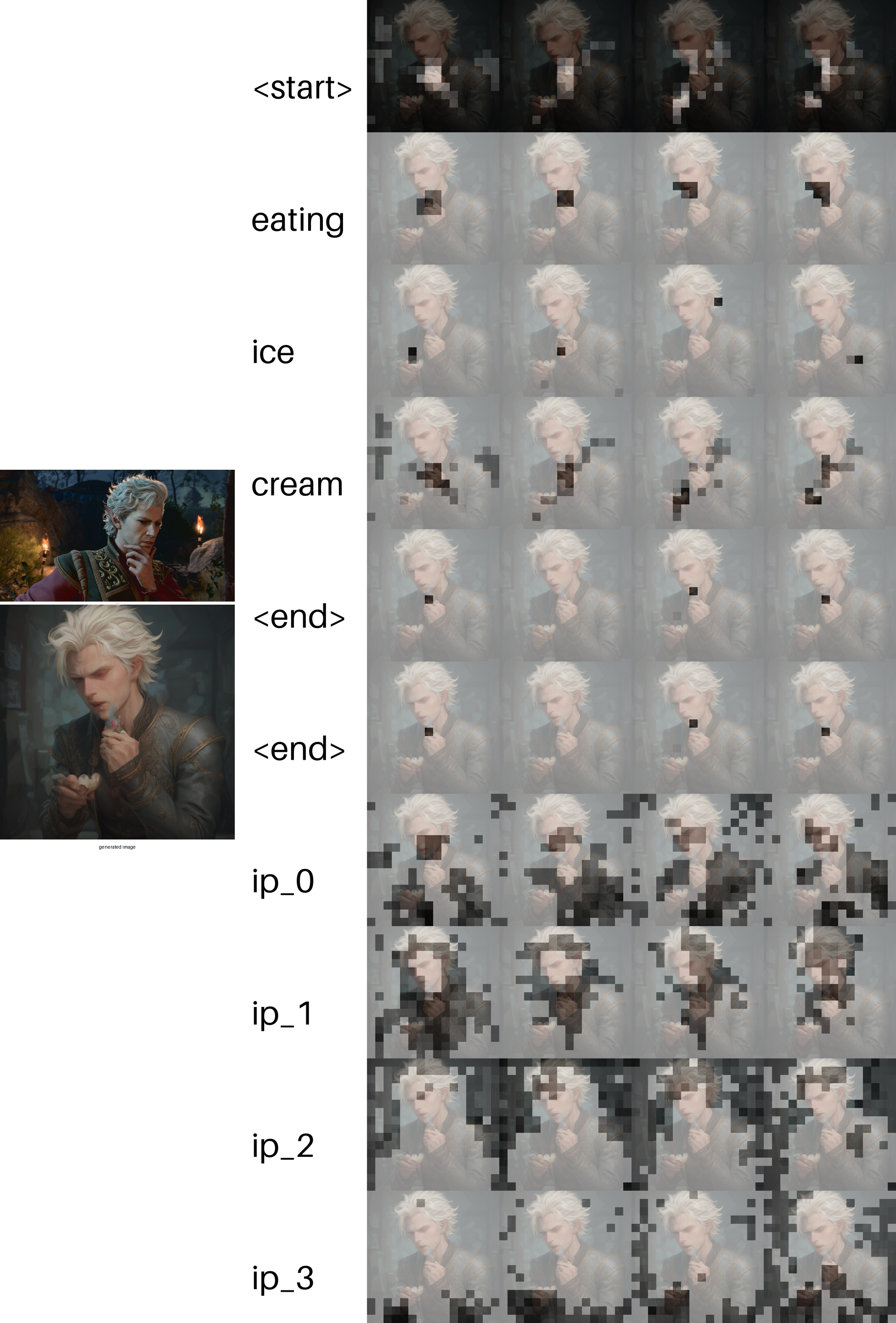}
    \caption{first transformer in the first up-block layer}
    \label{fig:map1}
\end{figure}

\newpage
\begin{figure}
    \centering
    \includegraphics[width=0.75\linewidth]{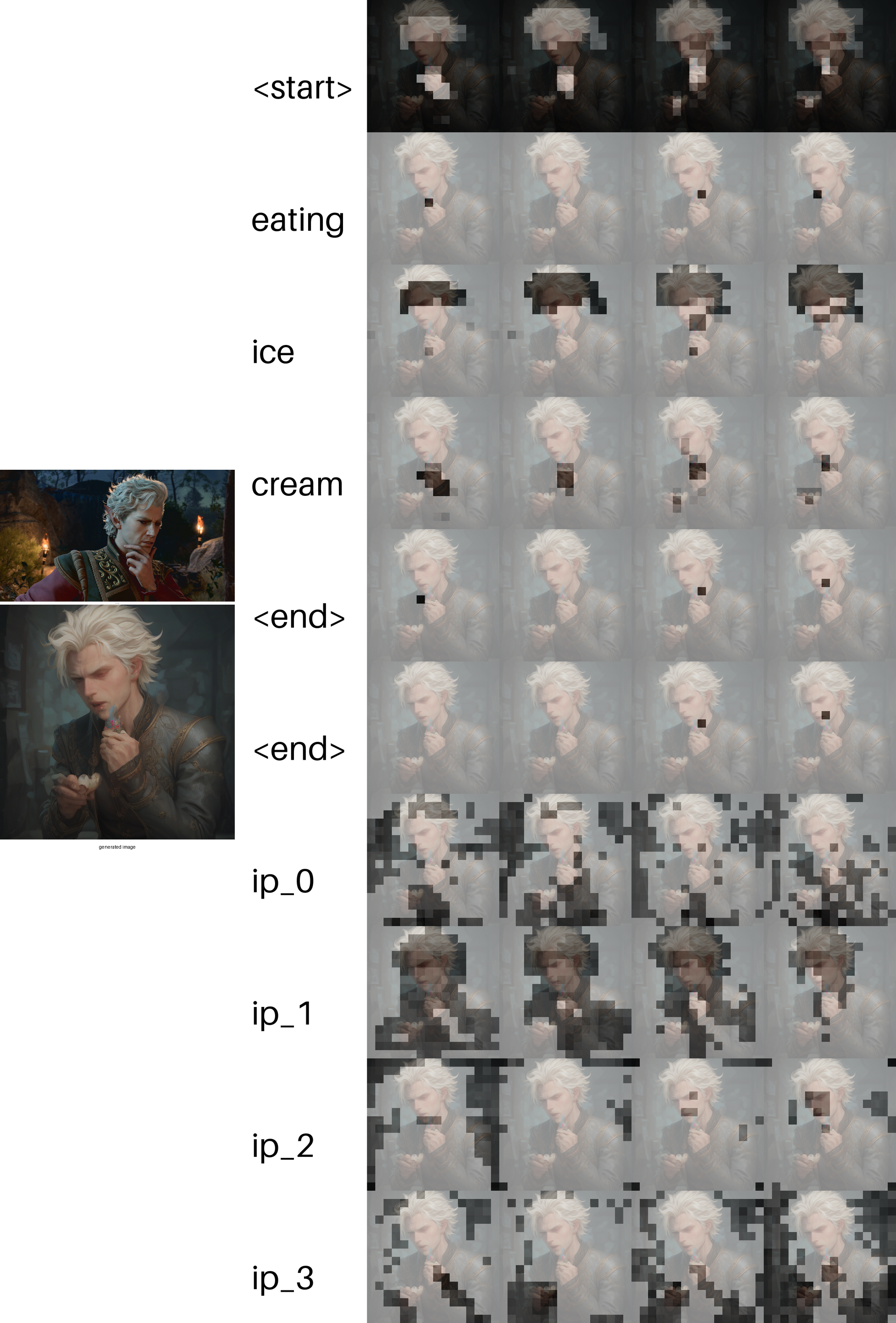}
    \caption{second transformer in the first up-block layer}
    \label{fig:map2}
\end{figure}

\newpage
\begin{figure}
    \centering
    \includegraphics[width=0.75\linewidth]{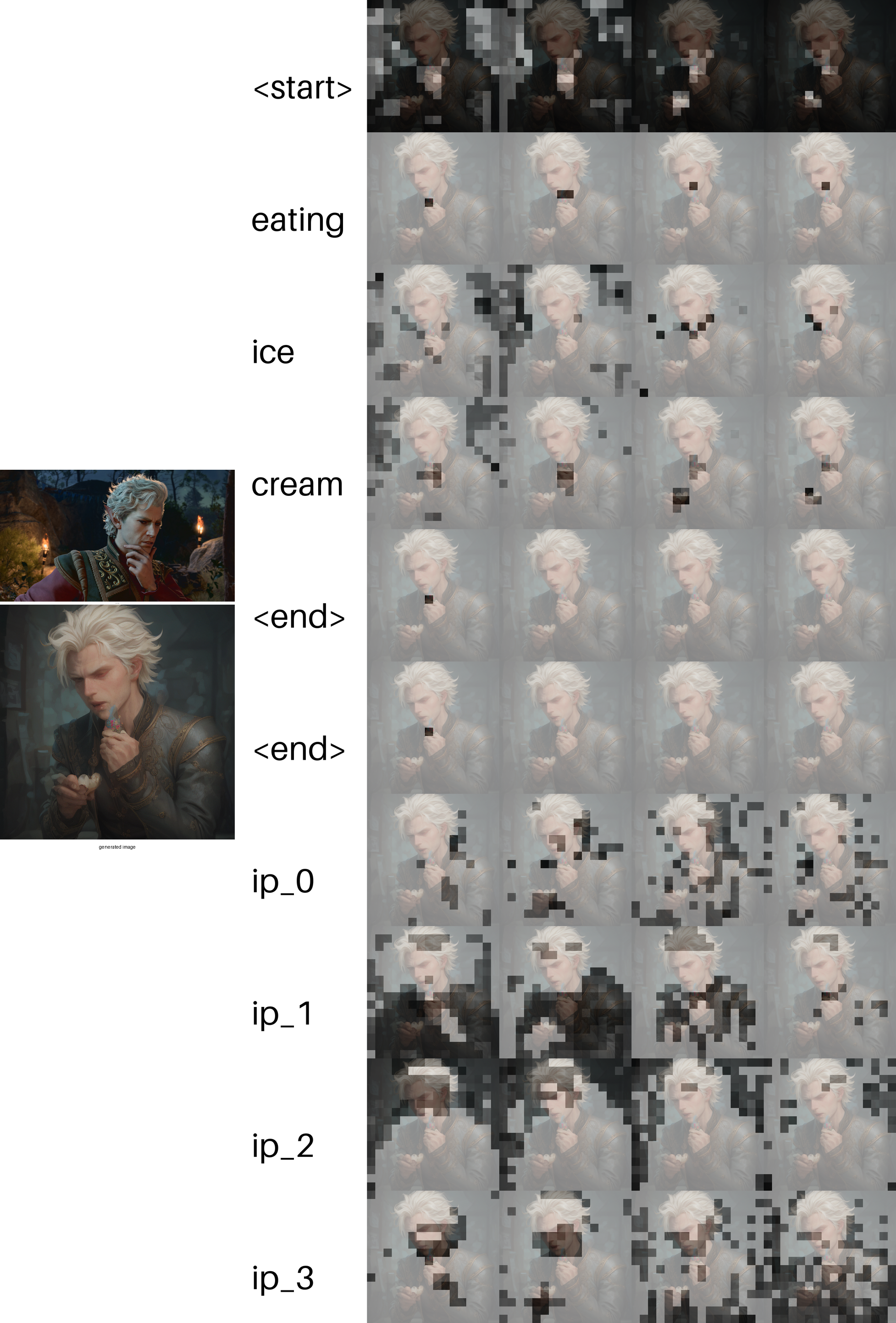}
    \caption{first transformer in the second up-block layer}
    \label{fig:map3}
\end{figure}

\newpage
\begin{figure}
    \centering
    \includegraphics[width=0.75\linewidth]{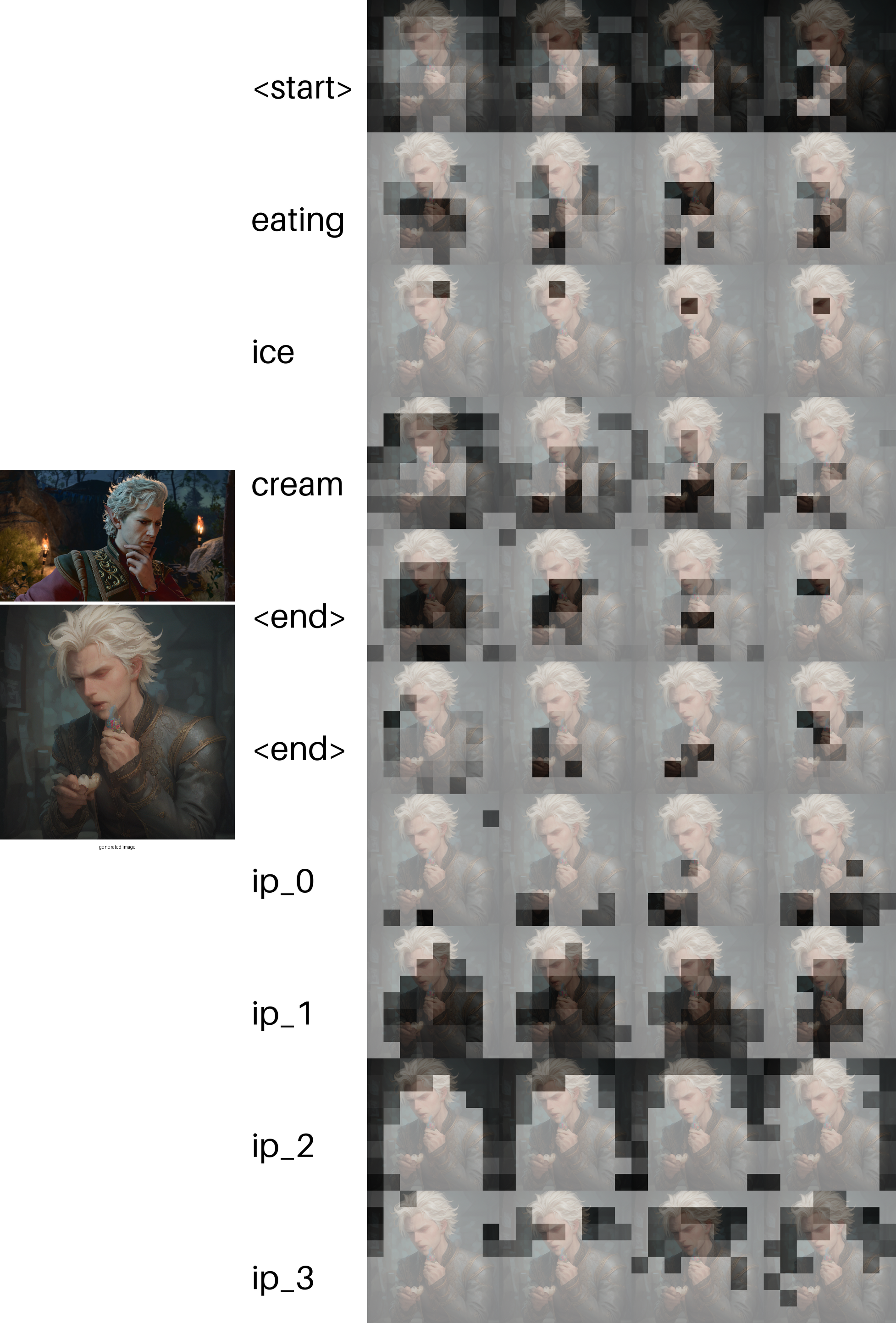}
    \caption{first transformer in the mid-block layer}
    \label{fig:placeholder}
\end{figure}
\end{document}

%% file: math_commands.tex
\usepackage{amsmath,amsfonts,bm}

\def\eqref#1{equation~\ref{#1}}

\def\1{\bm{1}}

\DeclareMathAlphabet{\mathsfit}{\encodingdefault}{\sfdefault}{m}{sl}
\SetMathAlphabet{\mathsfit}{bold}{\encodingdefault}{\sfdefault}{bx}{n}

